\documentclass{article}
\PassOptionsToPackage{hyphens}{url}
\usepackage{iclr2027_conference,times}
\usepackage{iftex}
\ifXeTeX
  \usepackage{fontspec}
\fi
\usepackage{amsmath,amssymb,booktabs,graphicx,xcolor,algorithm,algpseudocode,hyperref,url}
\usepackage{placeins,float}
\usepackage{longtable}
\let\cite\citep
\hypersetup{colorlinks=true,linkcolor=black,citecolor=black,urlcolor=blue}
\newcommand{\method}{Settle}
\newcommand{\stoptag}{\texttt{</think>}}
\newcommand{\KL}{\mathrm{KL}}
\newcommand{\ind}{\mathbf{1}}
\iclrfinalcopy
\makeatletter
\let\tableinput\@@input
\makeatother

\title{Settle: Learning When to Stop Reasoning}
\author{Ryan Brown$^{1}$ \quad Zihao Fu$^{2}$ \quad Chris Russell$^{1}$\\$^{1}$Oxford Internet Institute, University of Oxford\\$^{2}$Department of Linguistics and Modern Languages,\\The Chinese University of Hong Kong}
\begin{document}
\maketitle
\lhead{Preprint}

\begin{abstract}
Reasoning models often continue generating after their answers have settled. Settle learns when to stop from answer stability in completed traces. It trains the existing end-of-reasoning token while keeping other predictions close to the base model, and requires only ordinary decoding at inference. On MATH-500 with Qwen3-4B, Settle reduces token count by 40\% with a 0.5-percentage-point decrease in accuracy. It gains 6.16 percentage points over supervised fine-tuning on the same traces shortened at their first stable answer, at nearly identical token counts. Its stopping score predicts whether a correct answer will remain correct. Settle extends the accuracy--token-count Pareto frontier of the evaluated stopping methods.
\end{abstract}

\section{Introduction}
Maximally performant models such as Qwen3, Nemotron, and DeepSeek-R1-Distill use intermediate reasoning to improve answer quality. This reasoning can also consume computation checking and restating an answer that no longer changes. An end-of-reasoning token marks the transition between these two stages. We call the recorded response, including its reasoning and final answer, a \emph{trace}. Learning when to move from thinking to answering could save this computation while allowing difficult problems more reasoning.

\begin{figure}[t]
\centering
\includegraphics[width=\linewidth]{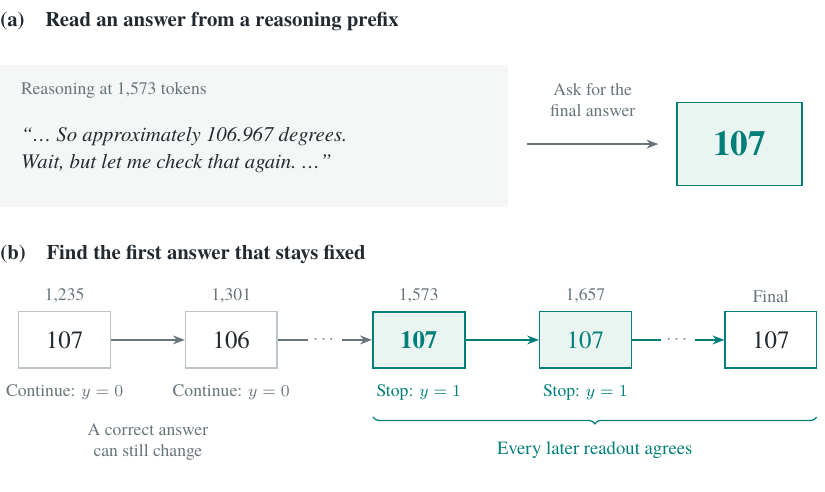}
\caption{\textbf{From answer readouts to stopping targets.} A recorded Qwen3-4B trace solves a pie-chart problem whose answer is $107^\circ$. (a) Prompting a saved prefix gives a short answer without changing the recorded continuation. (b) A first-correct-answer rule would stop at 1,235 tokens, but that answer changes later. Settle instead labels the stable suffix: from 1,573 tokens onward, every sampled answer agrees with the final readout. These boundaries receive stop targets; earlier boundaries receive continue targets. Ellipses mark omitted readouts; the final readout defines agreement and is not a training target.}
\label{fig:method}
\end{figure}

A trace reveals earlier opportunities to make that transition. We take successively longer initial portions of the reasoning, called \emph{prefixes}, and prompt the model to give a short answer from each one. Following \citet{liu_answer_2025}, an answer is \emph{stable} when it agrees with every later prompted answer, including the one prompted at the end of the recorded reasoning. A correct answer can still be unstable: in Figure~\ref{fig:method}, the answer reaches 107, changes to 106, and later returns to 107. These labels use the completed trace during training; the trained model decides from the reasoning generated so far.

Prior work reduces reasoning through supervised fine-tuning (SFT) on the shortest correct sampled solutions \cite{munkhbat_self-training_2025} or a separate controller trained to interrupt a frozen model \cite{liu_answer_2025}. Our stability-truncated SFT baseline trains on the same traces shortened at their first stable answer. Settle supervises the language model's own stopping decision, while regularizing its other predictions toward the base model.

We call this method \method{}: it learns to stop when the model has settled on an answer, rather than at the first answer that happens to be correct. We train this distinction directly into the model's existing end-of-reasoning token, with stop targets at stable prefixes and continue targets at earlier ones. A Kullback--Leibler (KL) penalty keeps predictions at other positions close to the base model. The labels depend only on the model's own answers. Answer checks take place before training; at inference, the model makes the stopping decision through standard decoding.

On Qwen3-4B/MATH-500, Settle gains 6.16 natural-accuracy points over stability-truncated SFT at nearly identical average token counts. It also beats the regularized SFT baseline by 1.42 points while using 25.2\% fewer tokens. Simply lowering the base model's token cap is less effective: Settle gains 7.03 forced-accuracy points over a 4,096-token cap while using 6.8\% fewer tokens. These results distinguish learning when to stop from imitating shorter solutions or imposing a smaller budget. Settle extends the Pareto frontier among the evaluated stopping methods, with savings extending to Qwen3-8B and a separate MATH test sample. Its closing-token score also predicts whether a correct answer will persist at identical prefix lengths, where the base score is near chance. All stopping decisions use ordinary decoding.

\section{Learning to stop from answer stability}
\label{sec:method}
Settle first labels the stable suffix of each recorded trace, then trains the model to recognize those stopping opportunities from its prefixes. Figure~\ref{fig:method} shows the label construction on a real example.

Let $x$ be a sampled response. A prefix $x_{<t}$ consists of its first $t$ generated tokens, conditioned throughout on the original question. We choose candidate boundaries $t_1<\cdots<t_K$ at paragraph breaks and at sentence endings before reflection phrases such as \emph{Wait} or \emph{Alternatively}. Appendix~\ref{app:implementation} specifies the spacing rule and the limit of 40 sampled boundaries per trace. The last prefix ends at $T$, immediately before the model's first end-of-reasoning tag (\stoptag{}), or at the end of the available response if no such tag occurs.

From a saved copy of each sampled prefix and of $T$, we append a closing tag and a final-answer prompt and greedily decode up to 16 tokens, following budget forcing \cite{muennighoff_s1_2025}. These prompted answers are the \emph{readouts} $a_1,\ldots,a_K,a_T$, with $a_T$ the \emph{terminal readout}. Each readout leaves the recorded continuation unchanged.

\begin{samepage}
An answer is \emph{stable at $t_j$} if every readout from $t_j$ onward agrees with the terminal readout $a_T$. Let $E(a,b)\in\{0,1\}$ indicate mathematical equivalence: normalized strings are compared first, followed by symbolic equivalence checking with Math-Verify \cite{kydlicek_math-verify_2025}; a failed comparison returns zero. Following \citet{liu_answer_2025}, the stop target is
\begin{equation}
 y_j=\prod_{k=j}^{K}E(a_k,a_T),\qquad j=1,\ldots,K.
 \label{eq:targets}
\end{equation}
\end{samepage}
Thus $y_j=1$ exactly when $E(a_k,a_T)=1$ for every $k\geq j$. If any later readout differs, $y_j=0$, even when $a_j=a_T$. The terminal readout serves as the comparison answer; only earlier boundaries receive training targets. Because the criterion compares the model's own answers, a consistently wrong answer also forms a stable suffix. Stopping is desirable here too: the remaining recorded reasoning consumes tokens without correcting the prompted answer. Evaluation uses fresh responses generated by the trained model.

Algorithm~\ref{alg:settle} in Appendix~\ref{app:implementation} computes the labels with $K$ terminal-answer comparisons and a backward scan. Answer readouts are generated once, offline, and the labelled traces are reused across training epochs. Deployment requires only the model's prediction from the current prefix.

For models with a single-token closing tag, define $p_\theta(t)=\pi_\theta(\stoptag{}\mid x_{<t})$. Let $\mathcal B(x)$ be the retained, non-terminal training boundaries and $\mathcal R(x)$ the response positions. We minimize
\begin{align}
 \mathcal L(\theta)={}&-\sum_x\sum_{j:t_j\in\mathcal B(x)}
       \left[w y_j\log p_\theta(t_j)+(1-y_j)\log(1-p_\theta(t_j))\right] \nonumber\\
 &+\lambda\sum_x\sum_{t\in\mathcal R(x)\setminus\mathcal B(x)}
       \KL\!\left(\pi_0(\cdot\mid x_{<t})\,\|\,\pi_\theta(\cdot\mid x_{<t})\right).
 \label{eq:loss}
\end{align}
The frozen reference $\pi_0$ is the base model. Increasing $w$ raises the penalty for continuing after stability, while $\lambda$ controls the KL penalty on other predictions. Both losses are summed over their positions. A trace therefore contributes one binary target per retained boundary and one reference-distribution target at each remaining response position.

At inference, the closing token competes with the other tokens in the model's output distribution. Sampling it ends reasoning and begins the final answer. This uses the same decoding procedure as the base model. Appendix~\ref{app:implementation} gives the boundary-loss derivative.

Full-sequence SFT learns when to stop as part of predicting each token. For a target continuation token $z\ne\stoptag{}$, let $q_\theta(z\mid x_{<t})$ be its probability conditional on not stopping. Its SFT loss decomposes as
\begin{equation}
\ell_{\mathrm{SFT}}(z,t)=-\log\pi_\theta(z\mid x_{<t})=-\log(1-p_\theta(t))-\log q_\theta(z\mid x_{<t}).
\label{eq:sft-decomposition}
\end{equation}
The first term teaches the model to continue; the second teaches which continuation token to generate. At the closing tag, SFT contributes $-\log p_\theta(t)$ and teaches termination. SFT thus learns a stopping point together with a particular sequence of reasoning and answer tokens.

Settle assigns a binary decision target at each labelled boundary. Before stability, the target is to continue; throughout the stable suffix, the target is to stop. This supplies several acceptable stopping opportunities from one trace. At other positions, the KL term uses the base model's full next-token distribution as the target. Labelled boundaries are excluded from that term so their closing-token probabilities can respond directly to the stop/continue labels. The experiments below compare this objective with sequence imitation and vary reference regularization within both procedures.

\section{Experimental design}
\label{sec:protocol}
We study Qwen3-4B, Qwen3-8B \cite{yang_qwen3_2025}, Llama-3.1-Nemotron-Nano-8B-v1 \cite{nvidia_llama-31-nemotron-nano-8b-v1_2025}, and DeepSeek-R1-Distill-Qwen-7B \cite{deepseek-ai_deepseek-r1_2025}. Rollouts are sampled at temperature 1.0 and top-$p=1.0$ from 50 DAPO-Math-17K problems \cite{yu_dapo_2025} selected for intermediate difficulty. Each model is trained on its own readouts. Known correct answers are used to select training problems and measure development accuracy. Stopping targets are constructed by comparing the model's answers. Unless noted, Qwen3-4B results average three independently trained models; Appendix~\ref{app:seeds} identifies the runs used throughout the evaluations.

We fit rank-32 attention-projection LoRA adapters \cite{hu_lora_2021} for three epochs with AdamW at $3\times10^{-4}$, then merge them into the model weights. Qwen3-4B uses 1,000 rollouts and 40,000 nonterminal readouts; one training run takes approximately 56 minutes on an H100. The main models use reference weight $\lambda=0.2$, with stopping weights $w\in\{0.10,0.25\}$. Appendix~\ref{app:implementation} gives the optimizer, label counts, and training context.

MATH-500 is the main benchmark for both Qwen3 sizes, using the same grading and overlap exclusions. We also evaluate the same Qwen3-4B models and stopping weight on GSM8K, OlympiadBench, GPQA-Diamond, AMC23, and Minerva. Appendix~\ref{app:transfer} gives sample counts and scoring details.

Evaluation uses thinking-mode chat templates, temperature 0.6, top-$p=0.95$, and caps of 8,192 or 16,384 tokens. MATH-500 \cite{hendrycks_math_2021} accuracy averages four samples on each of 498 problems, excluding two training overlaps; token means include all 500.

We report two accuracy measures. \emph{Natural accuracy} grades the final answer produced during ordinary generation, counting an unclosed reasoning block as wrong. \emph{Forced accuracy} additionally prompts responses that reach the token cap to give a final answer; responses that finish naturally are unchanged. Unless otherwise specified, token counts include these additional answer tokens. DEER's count also includes its trial answers, their short prompts, and generated tokens omitted from the final response. Input processing is excluded.

We select $w=0.10$ from $\{0.10,0.25\}$ by average natural accuracy on a 144-problem development set. We use this weight for the other models as well. A separate 500-problem sample from the MATH test set provides an additional evaluation; it is disjoint from the training set, development set, and MATH-500 (Appendix~\ref{app:chronology}).

Main base runs use sampling seeds 42--44; trained runs pair training seeds 0--2 with sampling seeds 42--44. We average samples and the three runs within each problem, then bootstrap whole problems with 20,000 draws. The intervals describe variation across evaluation problems for these models. Per-seed rows appear in Appendix~\ref{app:seeds}. The supplementary suite uses three trained models and one base run.

\section{Results}
\label{sec:results}
\begin{figure}[H]
\centering
\includegraphics[width=0.90\linewidth]{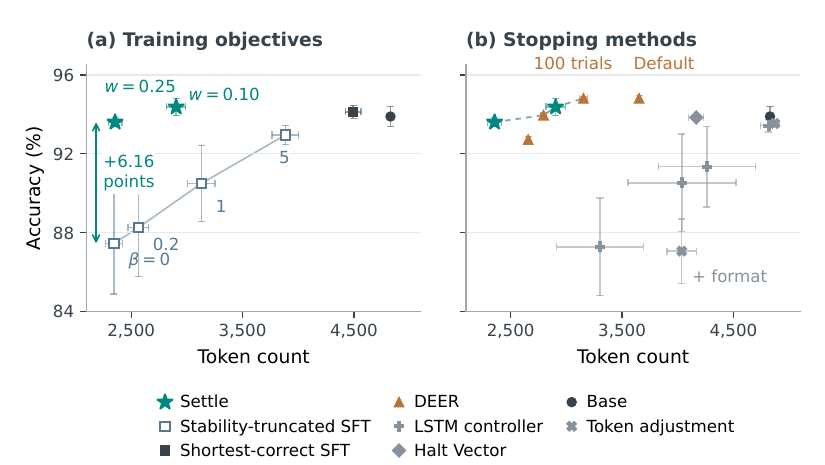}
\caption{\textbf{Settle improves on imitation and extends the Pareto frontier.} Qwen3-4B, MATH-500, 16k; three-run means $\pm$ sample standard deviations. (a) $\beta$ controls SFT regularization. (b) The dashed line joins nondominated means. Natural accuracy; token counts include DEER checks. Settle uses $\lambda=0.2$. Baselines: shortest-correct SFT \cite{munkhbat_self-training_2025}, LSTM and token adjustment \cite{liu_answer_2025}, Halt Vector \cite{jayabahu_halt_2026}, and DEER \cite{yang_dynamic_2025}. Settings and scores appear in Appendix~\ref{app:baselines}.}
\label{fig:comparison}
\end{figure}

\begin{samepage}
\subsection{Higher accuracy than imitating shorter solutions}
\label{sec:controls}
Our first baseline, \emph{stability-truncated SFT}, imitates Settle's source traces cut at their first stable answer. We append the answer readout and train every response token with cross-entropy, using all eligible traces, including incorrect ones. We compare this baseline with and without reference regularization; Settle uses $\lambda=0.2$ unless varied explicitly. SFT training settings are selected on development data (Appendix~\ref{app:controls}).

\end{samepage}

Both methods generate about 2,350 tokens per response, but Settle is substantially more accurate (Figure~\ref{fig:comparison}a). Settle at $w=0.25$ reaches 93.61\% natural accuracy, compared with 87.45\% for stability-truncated SFT without regularization: a gain of 6.16 percentage points, with paired 95\% interval $[4.70,7.68]$. Their exact token counts are 2,355 and 2,345, a difference of just 0.4\%. The gain remains at 6.17 points when unfinished responses are prompted to answer.

\emph{Shortest-correct SFT} uses a different kind of demonstration: the shortest complete correct solution sampled for each problem \cite{munkhbat_self-training_2025}. It reaches 94.13\% accuracy at 4,493 tokens. This baseline tests the benefit of selecting concise, correct solutions. Stability-truncated SFT tests the benefit of shortening Settle's source traces using the same answer-stability labels.

\begin{table}[t]
\centering
\caption{\textbf{Reference regularization within each training procedure.} MATH-500, Qwen3-4B, 16k; three-run means $\pm$ sample standard deviations. Accuracy grades naturally generated final answers; token counts cover ordinary generation. SFT uses mean losses and Settle uses sums, so $\beta$ and $\lambda$ have different scales. Other settings stay fixed within each procedure (Appendix~\ref{app:controls}).}
\label{tab:anchoring}
\small
\setlength{\tabcolsep}{6pt}
\begin{tabular}{llrr}
\toprule
Training procedure & Reference weight & Accuracy (\%) & Token count\\
\midrule
\tableinput{tables/anchor-main.tex}
\bottomrule
\end{tabular}
\end{table}

Adding a KL penalty strengthens stability-truncated SFT by keeping its predictions close to the base model (Table~\ref{tab:anchoring}). At the largest tested penalty, $\beta=5$, natural accuracy rises from 87.45\% to 92.96\%, while responses lengthen from 2,345 to 3,884 tokens.

Settle improves both accuracy and token count over this stronger SFT baseline. At $w=0.10$, it reaches 94.38\% accuracy with 25.2\% fewer tokens, a paired gain of 1.42 points $[0.42,2.49]$. The shorter Settle policy, $w=0.25$, uses 39.4\% fewer tokens and reaches 93.61\% accuracy; its mean accuracy difference is $+0.65$ points $[-0.54,+1.84]$.

Removing Settle's KL penalty produces repeated closing tags or missing final answers: natural accuracy falls to 0.03\%, and prompting unfinished responses to answer raises it only to 1.59\%. The penalty preserves answer generation while the binary loss trains termination. Appendix~\ref{app:controls} reports the full comparisons, including forced accuracy, paired intervals, and SFT restricted to the closing tag and final answer.

The savings persist across an eightfold range of positive KL weights (Table~\ref{tab:lambda-sensitivity}). Holding the stopping weight at $w=0.25$, we train three models for each $\lambda\in\{0.05,0.10,0.20,0.40\}$. All four settings use 44.2--57.4\% fewer tokens than the base model, with natural accuracy between 92.99\% and 93.98\%.

Stronger regularization produces longer responses, with mean token count rising from 2,057 to 2,695 across this sweep. Relative to $\lambda=0.20$, reducing the weight to 0.05 saves a further 12.0\% of tokens $[10.50,13.52]$, with an accuracy difference of $-0.15$ points $[-0.75,+0.45]$. Increasing it to 0.40 raises accuracy by 0.79 points $[0.08,1.49]$ and token count by 15.3\%. Thus the reference weight adjusts the accuracy--length tradeoff as well as preserving answer generation.

\begin{table}[t]
\centering
\caption{\textbf{Sensitivity to Settle's reference weight.} MATH-500, Qwen3-4B, 16k; $w=0.25$ and all other training settings fixed. Means $\pm$ sample standard deviations across three fits per coefficient. The sweep includes three new runs at the default $\lambda=0.20$. Token counts cover ordinary generation.}
\label{tab:lambda-sensitivity}
\small
\setlength{\tabcolsep}{7pt}
\begin{tabular}{rrrr}
\toprule
$\lambda$ & Natural accuracy (\%) & Forced accuracy (\%) & Token count\\
\midrule
\tableinput{tables/lambda-sensitivity.tex}
\bottomrule
\end{tabular}
\end{table}

Increasing the stopping weight $w$ raises the penalty for continuing after the answer stabilizes and shortens responses. Raising it from 0.10 to 0.25 reduces mean token count from 2,904 to 2,356, including prompted answers for unfinished responses. Forced accuracy changes from 94.73\% to 93.71\%. Reference-answer targets give nearby results; they differ from agreement labels at only 689 of 40,000 sampled boundaries (Appendix~\ref{app:controls}).

\subsection{Extending the frontier with ordinary decoding}
Figure~\ref{fig:comparison}b compares Settle with learned controllers, closing-token adjustments, and DEER's online answer checks. Both Settle settings are Pareto-optimal on Qwen3-4B/MATH-500: no compared method improves mean accuracy without using more tokens, or uses fewer tokens without lowering mean accuracy. We report natural final-answer accuracy averaged over three runs and count the tokens used by DEER's intermediate checks.

A long short-term memory (LSTM) controller trained on the same labels with the language model frozen \cite{liu_answer_2025} reaches 90.53\% accuracy at 4,038 tokens. Settle at $w=0.10$ gains 3.85 points $[2.76,5.00]$ with 28.1\% fewer tokens. Appendix~\ref{app:baselines} gives the development-selected threshold and full sweep.

Halt Vector moves a stopping intervention into the model weights \cite{jayabahu_halt_2026}; our Qwen3-4B adaptation reaches 93.84\% accuracy at 4,165 tokens. Think Token Adjustment changes the closing-token probability during inference \cite{liu_answer_2025}. It reaches 93.54\% at 4,872 tokens, or 87.06\% at 4,036 with the released answer-format constraints. Settle has higher natural accuracy and uses fewer tokens than both.

PUMA selects exits using semantic redundancy and answer checks \cite{min_stop_2026}. In a single-seed comparison on the same Qwen3-4B traces, PUMA-SFT reaches 93.07\% at 2,803 tokens; Settle at $w=0.25$ reaches 93.93\% with 16.8\% fewer tokens. Released PUMA replay reaches 93.42\% at 3,433 tokens including checks (Appendix~\ref{app:puma}).

DEER uses confidence in trial answers to decide when to exit \cite{yang_dynamic_2025}. These checks require extra generation even when reasoning continues; Settle generates no trial answers at inference. DEER's default and 100-trial settings both reach 94.81\% accuracy, at 3,653 and 3,153 tokens. Tuning DEER on development data under the two Settle token budgets gives 93.96\% at 2,795 tokens and 92.72\% at 2,658 tokens (Appendix~\ref{app:baselines}).

Settle extends this frontier. At $w=0.25$, it uses 11.4\% fewer tokens than the 2,658-token DEER setting, with an accuracy difference of $+0.89$ points $[-0.03,+1.86]$. Against the 2,795-token setting, it saves 15.7\% with a difference of $-0.35$ points $[-1.07,+0.38]$. Section~\ref{sec:breadth} compares inference times.

\subsection{Higher accuracy than a smaller token budget}
Settle reduces token counts across Qwen3 sizes and inference budgets (Table~\ref{tab:main}). We compare these budgets using forced accuracy (Section~\ref{sec:protocol}). At a 16k cap, Qwen3-4B uses 39.9\% fewer tokens, from 4,829 to 2,904, with an accuracy decrease of 0.50 points. At 8k, it saves 36.5\% with a decrease of 0.13 points. Applying the same training rule and stopping weight to Qwen3-8B saves 38.7\%, from 5,112 to 3,135 tokens, with a decrease of 0.22 points.

\begin{table}[t]
\centering
\caption{\textbf{MATH-500 results with agreement labels, $w=0.10$.} Forced accuracy in percent; changes in points with paired 95\% intervals. Tokens are base / Settle means, including additional answer tokens. Each row averages three runs per model. ``8k'' and ``16k'' denote 8,192 and 16,384.}
\label{tab:main}
\footnotesize
\setlength{\tabcolsep}{3pt}
\begin{tabular}{llrrlrr}
\toprule
Model & Cap & Base & Settle & Change [95\% interval] & Token count & Saved \%\\
\midrule
\tableinput{tables/main.tex}
\bottomrule
\end{tabular}
\end{table}

Learning when to stop outperforms simply lowering the base model's token cap. Across six tested caps (Figure~\ref{fig:main}, Appendix~\ref{app:controls}), the 4,096-token cap gives 87.70\% forced accuracy at 3,115 tokens on average. Settle at $w=0.10$ reaches 94.73\% at 2,904 tokens: 7.03 points higher accuracy with 6.8\% fewer tokens. It also has higher accuracy and lower mean token count than the 6,144- and 8,192-token-cap runs.

The difference is larger for shorter responses. A 2,560-token cap gives 80.77\% accuracy at 2,314 tokens, compared with 93.71\% at 2,356 for Settle ($w=0.25$). The 16k base model retains the highest forced accuracy, but reducing its cap has a substantially steeper accuracy cost than learned stopping.

Earlier stopping can also improve the chance of producing a final answer within the budget. For Qwen3-4B at 16k, natural accuracy rises from 93.89\% to 94.38\%, while forced accuracy changes from 95.23\% to 94.73\%. Thus the natural-accuracy gain reflects improved completion within the budget; it accompanies a small decrease under forced scoring.

\subsection{Answer stability and the learned stopping score}
\label{sec:readouts}
\begin{figure}[t]
\centering
\includegraphics[width=\linewidth]{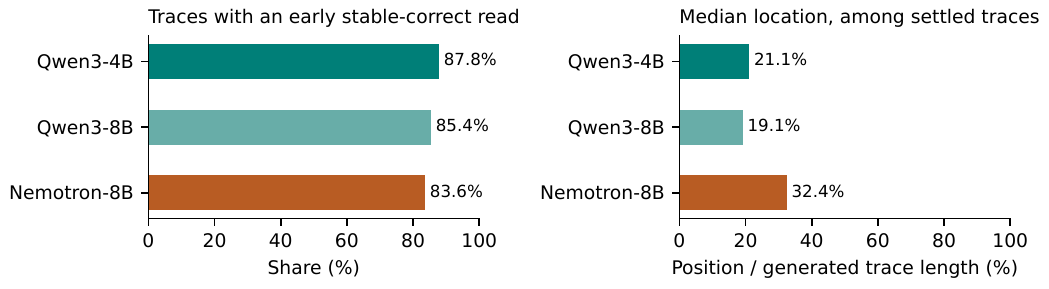}
\caption{\textbf{Stable correct answers often precede completion.} MATH-500; 500 separately sampled traces per model, with an 8,192-token cap. Left: fraction with an early stable correct answer. Right: median first stable position as a fraction of trace length, among those traces. Appendix~\ref{app:census} gives full results and a repeated Qwen3-4B sample.}
\label{fig:census}
\end{figure}

The base models often reach a stable correct answer well before completing their responses (Figure~\ref{fig:census}). On MATH-500, 87.8\% of Qwen3-4B traces have an early correct answer that persists at every later readout. Among these traces, the first such readout occurs at a median of 21.1\% of the generated length. Qwen3-8B and Nemotron show similar patterns. This diagnostic uses reference answers to assess correctness; the training labels require only agreement between readouts.

We next test whether Settle's stopping score predicts answer stability beyond the amount of reasoning already generated. We compare the base and trained models' closing-token probabilities on the same prefixes, alongside predictors that use only token count. The target is \emph{stable correctness}: the current answer and every later readout must be correct. Reference answers define this diagnostic target; the training labels use only agreement between the model's own answers.

The score is evaluated before the next token is generated. Later readouts determine the evaluation target but are not supplied to the predictor. Scoring every model on the same recorded prefixes holds the reasoning text fixed while testing what the closing-token probability encodes. The generation experiments separately measure the accuracy and length of each model's own responses.

We compare prefixes at identical token counts, pairing a stably correct answer with one that is not. A second comparison considers only prefixes that already give the correct answer, testing whether the score predicts which answers will stay correct. We measure discrimination by the area under the receiver operating characteristic curve (AUROC); 0.5 indicates chance performance.

\begin{table}[t]
\centering
\caption{\textbf{Predicting answer persistence beyond elapsed length and present correctness.} Overall and conditional AUROC [95\% interval] on fixed Qwen3-4B prefixes. Overall uses 17,899 boundaries; matching token counts uses 12,106. The final column also requires the present readout to be correct, leaving 2,455 eligible boundaries. Intervals resample whole traces; each weight uses one agreement-trained model. $\dagger$ A token-count predictor assigns the same score to equal-length prefixes, giving conditional AUROC 0.5. Appendix~\ref{app:readouts} defines the estimands and reports all matching ranges.}
\label{tab:readouts}
\small
\setlength{\tabcolsep}{3pt}
\begin{tabular}{lrrr}
\toprule
Score & Overall & \shortstack{Same token\\count} & \shortstack{Correct now,\\same token count}\\
\midrule
\tableinput{tables/native-position-main.tex}
\bottomrule
\end{tabular}
\end{table}

Across 17,899 boundaries from 498 MATH-500 traces, Settle reaches AUROCs of 0.831 and 0.838 for $w=0.10$ and $w=0.25$, respectively (Table~\ref{tab:readouts}). The base score reaches 0.532 and token-count predictors reach 0.618. Matching \emph{identical} token counts leaves both Settle scores above 0.80, while the base score falls to 0.507. A token-count predictor necessarily ties, giving AUROC 0.5. Wider matching ranges give consistent results (Appendix~\ref{app:readouts}).

Among currently correct answers at identical token counts, Settle predicts which stay correct at every later readout, with AUROCs of 0.695 and 0.706. The base score is near chance at 0.495. Paired improvements are 0.200 $[0.112,0.290]$ and 0.210 $[0.119,0.305]$, showing that the score predicts persistence beyond present correctness and elapsed reasoning.

Settle predicts stability through the same closing-token score it uses to end generation.

\subsection{Generalization and inference time}
\label{sec:breadth}
Settle saves 42.7\% of tokens on a separate 500-problem MATH sample, with forced accuracy of 96.52\% versus 96.33\% for the base model. The same three Qwen3-4B models at $w=0.10$ reduce mean token count from 3,803 to 2,179 at the 16k cap (Appendix~\ref{app:chronology}).

We keep the stopping weight fixed across five other benchmarks (Table~\ref{tab:transfer-main}). On GSM8K \cite{cobbe_training_2021}, Settle saves 43.2\% of tokens with a 0.68-point forced-accuracy decrease. On OlympiadBench \cite{he_olympiadbench_2024} and GPQA-Diamond \cite{rein_gpqa_2023}, token counts fall by 36.3\% and 13.2\%, while accuracy rises by 3.76 and 0.34 points. On AMC23 and Minerva, token savings of 35.6\% and 49.1\% accompany forced-accuracy decreases of 2.6 and 3.7 points. The fixed stopping weight therefore transfers token savings more consistently than accuracy (Appendix~\ref{app:transfer}).

\begin{table}[H]
\centering
\caption{\textbf{Token savings transfer across tasks.} Qwen3-4B, $w=0.10$, 16k cap. Forced accuracy and token pairs are base / Settle. All rows average three trained models; MATH uses three base runs and the other tasks one. Sample sizes and protocols appear in Appendices~\ref{app:chronology} and~\ref{app:transfer}.}
\label{tab:transfer-main}
\small
\setlength{\tabcolsep}{6pt}
\begin{tabular}{lrrr}
\toprule
Benchmark & Accuracy (\%) & Token count & Saved (\%)\\
\midrule
MATH, separate sample & 96.33 / 96.52 & 3,803 / 2,179 & 42.7\\
GSM8K & 95.1 / 94.5 & 2,053 / 1,167 & 43.2\\
OlympiadBench & 66.6 / 70.4 & 9,024 / 5,750 & 36.3\\
GPQA-Diamond & 55.6 / 55.9 & 8,565 / 7,433 & 13.2\\
AMC23 & 93.4 / 90.8 & 7,617 / 4,903 & 35.6\\
Minerva & 48.2 / 44.5 & 6,536 / 3,327 & 49.1\\
\bottomrule
\end{tabular}
\end{table}

On 100 MATH problems using one H100, batch-1 median latency falls from 13.25 to 7.66 seconds. At batch size 32, total time is 302 seconds for Settle, 324 for base, and 521 for DEER, with natural accuracies of 94\%, 94\%, and 95\%. The batch-32 saving over base is about 7\% of total time, showing that fewer generated tokens need not yield a proportional runtime reduction. Appendix~\ref{app:latency} reports the timing procedure and all batch sizes.

\section{Related work}
\label{sec:related}
\citet{liu_answer_2025} introduce the future-agreement labels used here and train an LSTM stopping predictor on hidden activations of a frozen language model. They also study online answer consistency and inference-time closing-token adjustment. Settle builds on their labels to train the existing closing-token distribution, which makes the language model itself the stopping policy. Online methods obtain evidence through additional computation: DEER uses intermediate answer confidence \cite{yang_dynamic_2025}, while SABER tests stability with branch probes \cite{cheng_saber_2026}. LYNX learns confidence-controlled exits \cite{akgul_lynx_2025}, and TERMINATOR adds a transformer layer and prediction head trained on first-answer positions \cite{nagle_terminator_2026}. Settle learns from offline answer agreement and uses the existing output distribution, without an extra inference module.

Budget forcing controls reasoning length directly \cite{muennighoff_s1_2025}. Training methods can instead change the generated policy: self-training imitates the shortest correct sampled solution \cite{munkhbat_self-training_2025}, and reinforcement-learning methods such as ThinkPrune optimize efficient reasoning \cite{hou_thinkprune_2025}. ReasonMaxxer applies sparse offline corrections with a reference anchor \cite{akgul_rethinking_2026}, motivating targeted policy updates without an online RL loop. Several methods also learn stopping behavior in the weights. Halt Vector internalizes a causal steering intervention in an attention-only adapter \cite{jayabahu_halt_2026}. ESTAR trains a separate stop-proposal token through SFT and reinforcement learning, with an external classifier accepting proposals during inference \cite{wang_estar_2026}. PUMA learns exit positions selected using semantic redundancy and answer verification through SFT, DPO, or GRPO \cite{min_stop_2026}. Settle combines binary supervision of the existing stop token at multiple agreement-labelled boundaries with reference regularization elsewhere. The SFT controls use the same traces, with and without a reference penalty, to compare this decision loss with sequence imitation.

\section{Limitations and future work}
\label{sec:limitations}
Our emphasis on mathematical reasoning follows prior work on concise generation \cite{munkhbat_self-training_2025}; additional model and benchmark evaluations test transfer. The calibration study motivates testing development-selected stopping weights across more tasks. Denser boundary sampling could improve stopping performance by locating the onset of stability more precisely. Longer rollouts and alternative readouts would extend label evidence; fresh on-policy traces could better match deployment prefixes.

Our regularization experiments hold other settings fixed. Jointly varying stopping pressure, regularization, supervision coverage, and training duration would map further accuracy--length tradeoffs. Interventions on the learned stop score could test its causal contribution. Further training runs and serving studies across hardware and batch sizes would clarify reproducibility and practical efficiency.

\section{Conclusion}
Settle learns when to stop from completed reasoning traces and acts through ordinary decoding. On Qwen3-4B/MATH-500, it saves 39.9\% of tokens versus the base model, with a 0.50-point decrease in forced accuracy. It gains 6.16 natural-accuracy points over stability-truncated SFT at nearly identical token counts, and beats regularized SFT by 1.42 points with 25.2\% fewer tokens. Against a 4,096-token cap, it gains 7.03 forced-accuracy points while using 6.8\% fewer tokens. These comparisons favour learning stopping over sequence imitation and fixed budgets. Savings extend to Qwen3-8B and a separate MATH sample, and persist across an eightfold range of KL weights. The learned score predicts whether correct answers will persist even at identical token counts. The stopping policy is encoded in the merged model weights. Deployment uses the base model's vocabulary and decoding procedure, with no extra stopping head or online answer checks. Together, these findings support answer stability as supervision for efficient reasoning through the model's existing stopping action, with an accuracy--length tradeoff that depends on the model and task.

\clearpage
\subsection*{Reproducibility statement}
Appendix~\ref{app:implementation} specifies boundary construction, training, model selection, scoring, and bootstrap estimation. Appendix~\ref{app:chronology} documents the separate MATH evaluation sample, and Appendix~\ref{app:seeds} reports individual runs. Appendix~\ref{app:readouts} describes the native-score diagnostics. The remaining appendices give the full benchmark results, training and truncation controls, baseline comparisons, and serving measurements, including hardware and batch settings.

\bibliography{references}
\bibliographystyle{iclr2027_conference}

\clearpage
\appendix
\raggedbottom
\section{Stopping supervision and reference regularization}
\label{app:controls}
The first stable answer can supply either a shorter demonstration or a target for the stopping decision. Our central comparison uses the same traces to test these two learning procedures. Settle improves accuracy over stability-truncated SFT at nearly identical token counts; the additional controls examine how reference regularization and the choice of stopping targets contribute to this result.

For stability-truncated SFT, we cut each source response at its first future-agreement boundary and append the closing tag and answer readout. Cross-entropy supervises the entire target, with question tokens masked, no reference anchor, and no correctness filter. Of 1,000 responses, 993 have an eligible boundary and 976 fit the 8,192-token context. Adapters match Settle's rank and scaling. We use AdamW with zero weight decay, accumulation over eight responses, gradient clipping at 1, and a linear schedule with 10\% warmup. Three seeds are trained at each learning rate, $10^{-4}$ and $3\times10^{-4}$; all three epochs are evaluated on 144 development problems using training seed 0.

To compare shorter policies, we select the highest development accuracy within 50\%, 60\%, or 75\% of the base model's token count, breaking ties by lower count. The 50\% budget selects epoch 1 at $3\times10^{-4}$; both larger budgets select epoch 2 at the same rate, which also has the highest accuracy overall. These choices are fixed for all three seeds before MATH-500 evaluation. Table~\ref{tab:agreement-sft} gives both durations, and epoch 2 enters the main comparison.

\begin{table}[!htbp]
\centering
\caption{\textbf{SFT on stability-truncated solutions.} MATH-500 at 16k, four responses per problem. Accuracy excludes the two overlapping problems; token means include all 500. All rows average three independently trained models. Epochs are selected on development data.}
\label{tab:agreement-sft}
\footnotesize
\begin{tabular}{lrrr}
\toprule
Training procedure & Natural \% & Forced \% & Token count\\
\midrule
\tableinput{tables/agreement-sft.tex}
\bottomrule
\end{tabular}
\end{table}

We vary the reference constraint within each learning procedure. Settle compares $\lambda=0$ with $\lambda=0.2$ at $w=0.25$, keeping all 1,000 source traces, boundary labels, supervised positions, and 375 updates fixed. Stability-truncated SFT uses the same 976 targets, with reference KL added to the retained reasoning prefix. Let $\mathcal C$ denote the shortened completion, including its closing tag and answer, and $\mathcal P\subset\mathcal C$ its reasoning prefix. The per-response loss is
\[
\mathcal L_{\mathrm{SFT+KL}}=
-\frac{1}{|\mathcal C|}\sum_{t\in\mathcal C}\log\pi_\theta(x_t\mid x_{<t})
+\frac{\beta}{|\mathcal P|}\sum_{t\in\mathcal P}
\KL\!\left(\pi_0(\cdot\mid x_{<t})\,\|\,\pi_\theta(\cdot\mid x_{<t})\right).
\]
The KL term acts on the reasoning prefix, excluding the question and appended ending. We test $\beta\in\{0,0.2,1.0,5.0\}$ with three seeds at the selected learning rate $3\times10^{-4}$. Training ends at the epoch-2 checkpoint, after 244 updates of the original three-epoch, 366-update schedule. The remaining optimizer and evaluation settings stay fixed. Because SFT averages each loss over positions whereas Settle sums them, the coefficients have different scales.

The two procedures use the same source traces and agreement labels; their losses, supervised positions, and training durations are specified above. Appendix~\ref{app:baselines} reports all scores.

The positive-weight sweep in Table~\ref{tab:lambda-sensitivity} uses twelve additional fits: three training seeds at each $\lambda\in\{0.05,0.10,0.20,0.40\}$, with $w=0.25$ throughout. All settings use the same 1,000 labelled traces, 8,192-token training context, 375 updates, adapters, optimizer, and evaluation protocol. The sweep includes three new training runs at the default $\lambda=0.20$. Figure~\ref{fig:comparison} and Table~\ref{tab:anchoring} retain the original models. Each fit and evaluation uses one H200. Table~\ref{tab:lambda-perseed} gives the individual runs.

Without KL regularization, 5,995 of 6,000 responses contain no extracted final answer. Table~\ref{tab:anchor-perseed} reports the individual runs.

\begin{table}[!htbp]
\centering
\caption{\textbf{Both Settle settings versus every SFT reference weight.} Settle-minus-SFT accuracy differences in percentage points, with paired 95\% problem-bootstrap intervals. Token savings compare ordinary-generation means; negative values mean Settle uses more tokens. All comparisons use three trained models per setting and the interval procedure in Appendix~\ref{app:implementation}.}
\label{tab:settle-sft-comparisons}
\small
\setlength{\tabcolsep}{6pt}
\begin{tabular}{lrrr}
\toprule
SFT $\beta$ & Natural accuracy change & Forced accuracy change & Tokens saved (\%)\\
\midrule
\tableinput{tables/settle-sft-comparisons.tex}
\bottomrule
\end{tabular}
\end{table}

Table~\ref{tab:controls} compares the supervision targets and stopping rules. Reference-answer and agreement supervision use the main source traces. Ending-only SFT, first-positive-only training, and shuffled labels use a separate trace sample; the latter two also use a higher stopping weight.

\begin{table}[h]
\centering
\caption{Training controls on MATH-500 at 16k. Accuracy uses the same scoring and overlap exclusions as the main evaluation. Tokens include additional answer tokens under forced scoring. Reference labels use known correct answers; agreement labels use the model's own answers.}
\label{tab:controls}
\footnotesize
\setlength{\tabcolsep}{7pt}
\begin{tabular}{lrrrr}
\toprule
Supervision & Runs & Natural \% & Forced \% & Token count\\
\midrule
Agreement, $w=0.10$ & 3 & 94.38 & 94.73 & 2,904\\
\tableinput{tables/controls.tex}
\bottomrule
\end{tabular}
\end{table}

\begin{figure}[t]
\centering
\includegraphics[width=\linewidth]{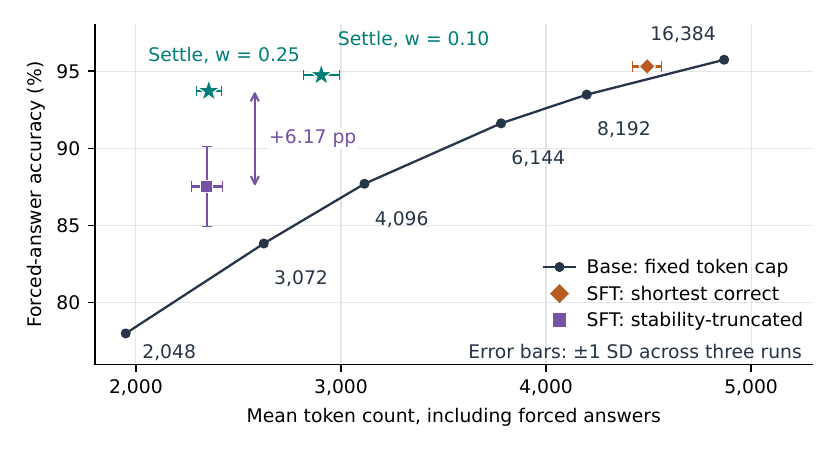}
\caption{\textbf{Learning the stop action versus shorter budgets and demonstrations.} Labels on the base curve give maximum token budgets. Every point uses forced-answer accuracy and generated-token count, with the same scoring and overlap exclusions. Each base cap uses one run; Settle and both SFT controls show three-run means with sample-standard-deviation bars at a 16k cap. SFT on stability-truncated solutions uses the development-selected setting; the arrow gives its forced-accuracy difference from Settle at $w=0.25$. The base curve uses a separate evaluation from Table~\ref{tab:main}; lines connect its evaluated caps.}
\label{fig:main}
\end{figure}

A uniform token cap shortens reasoning without changing the model. Figure~\ref{fig:main} compares this intervention with learned stopping; Table~\ref{tab:budget-curve} gives the six caps, using Qwen3-4B, seed 42, four responses per problem, and the common protocol. The additional 2,560-token cap gives 80.77\% forced accuracy at 2,314 mean tokens (Table~\ref{tab:controls}).

\begin{table}[h]
\centering
\caption{Qwen3-4B fixed-budget curve on MATH-500. Each row is one evaluation run.}
\label{tab:budget-curve}
\begin{tabular}{rrr}
\toprule
Maximum tokens & Forced accuracy \% & Mean token count\\
\midrule
\tableinput{tables/budget-curve.tex}
\bottomrule
\end{tabular}
\end{table}

Following \citet{munkhbat_self-training_2025}, shortest-correct SFT chooses the shortest complete correct response per training problem that fits the context. It imitates this response through its natural ending. This retains a complete solution, whereas stability-truncated SFT constructs a new ending at a labelled boundary.

Ending-only SFT retains each rollout through its first stable-correct nonterminal boundary; rollouts without one are excluded. Negative log-likelihood supervises the appended closing tag, final-answer prompt, forced answer, and end-of-response token, while reference KL with coefficient 0.2 constrains the retained prefix. Training uses the separate Qwen3-4B trace sample, an 8,192-token context, and seeds 0 and 1.

Table~\ref{tab:self-closure} uses the common evaluation protocol with seed 42 and four responses per problem: both models at 8k and training seed 0 at 16k.

\begin{table}[h]
\centering
\caption{Ending-only SFT on MATH-500. Accuracy is natural, with no forced answer at the cap.}
\label{tab:self-closure}
\begin{tabular}{rrrr}
\toprule
Maximum tokens & Training seed & Natural accuracy \% & Token count\\
\midrule
\tableinput{tables/self-closure.tex}
\bottomrule
\end{tabular}
\end{table}

We isolate the placement of positive stopping supervision by concentrating each trace's total positive loss weight at its first stable boundary. Later positive boundaries receive zero stopping weight, while the negative targets and all positions receiving KL regularization stay unchanged. Thus both methods have the same positive supervision weight per trace. Both use the original 1,000 traces, training seed 0, $w=0.10$, $\lambda=0.2$, and 375 updates.

\begin{table}[h]
\centering
\caption{Placement of positive stopping supervision, Qwen3-4B. Positive loss weight per trace, negative targets, and KL positions are identical. One training seed; four responses per MATH-500 problem, with evaluation seed 42 and a 16k cap. Token counts include prompted answers for capped responses.}
\label{tab:first-stable}
\small
\begin{tabular}{llrrr}
\toprule
Task & Positive supervision & Natural \% & Forced \% & Token count\\
\midrule
MATH-500 & First stable boundary & 93.42 & 93.93 & 2,969\\
 & All stable boundaries & 94.18 & 94.58 & 2,890\\
\bottomrule
\end{tabular}
\end{table}

On MATH-500, supervising all stable boundaries raises natural accuracy by 0.75 points, with paired 95\% problem-bootstrap interval $[-0.15,+1.71]$, and uses 2.6\% fewer tokens. This comparison holds supervision weight and regularization positions fixed.

A correct readout need not remain correct as reasoning continues. We test whether that future information improves stopping targets, using the same 1,000 traces, 39,649 positions, adapters, KL positions, and optimizer. Persistent labels require correctness at the current and all later readouts; current-only labels use the present readout. The current-only comparison uses both the same stopping weight and reweighted positive and negative losses that preserve their total contribution.

At 16k, natural accuracy is 94.03\% for persistent labels, 93.12\% for current-only labels, and 93.32\% after reweighting. Mean token counts are 2,406, 2,312, and 2,322; forced accuracy is 94.18\%, 93.26\%, and 93.46\%, respectively. Persistent labels perform best at 16k. At 8k the ranking varies by condition.

We also test a fixed stopping threshold applied to the learned closing probability. On the separate MATH sample, a threshold of 0.5 applied to the reference-answer-trained model at $w=0.25$ gives 94.67\% forced accuracy, versus 96.33\% for base (paired difference $-1.67$ points, 95\% interval $[-2.90,-0.45]$).
\FloatBarrier
\section{Comparisons with alternative stopping policies}
\label{app:baselines}
Both Settle settings form part of the accuracy--token-count Pareto frontier in Figure~\ref{fig:comparison}. Table~\ref{tab:learned-baselines} gives the complete comparison under the common protocol, including the tokens used by DEER's intermediate checks (Appendix~\ref{app:implementation}).

Following \citet{liu_answer_2025}, a one-layer LSTM with 128 hidden units predicts agreement from frozen final-layer activations at Settle's labelled prefixes. Validation F1, with problems separated between fitting and validation, selects training duration before refitting on the full pool with three seeds. The controller reads activations every 128 generated tokens in place of the original sentence schedule. At the first threshold crossing, it closes reasoning and generates a greedy answer of at most 256 tokens within the remaining budget. We evaluate by replaying traces, exposing each decision only to the available prefix activations. Token count includes the retained prefix and answer; the inserted answer prompt is counted separately.

We evaluate controller thresholds $0.5,0.9,0.95,0.99,0.995,0.999,0.9995,0.9999$ on 144 development problems with four responses each. The highest natural accuracy within 50\%, 60\%, and 75\% of base token count selects $\tau=0.95,0.99,0.995$, respectively, with ties broken by lower count. We also test $\tau=0.9999$, which has the highest development accuracy overall. These thresholds are fixed for all three trained controllers before MATH-500 evaluation; the main comparison uses the 60\% budget.

\begin{table}[t]
\centering
\caption{\textbf{Learning when to stop: training controls and existing stopping methods.} MATH-500, Qwen3-4B, 16k cap; three-run means $\pm$ sample standard deviations. Natural accuracy grades generated final answers; forced accuracy also prompts budget-limited responses to answer. Token counts cover ordinary generation before additional answer forcing, including intermediate checks for DEER. Bold marks the best mean within each panel. Unregularized SFT and controller settings are selected on development data; regularization controls keep other training settings fixed (Appendix~\ref{app:controls}).}
\label{tab:sft}
\label{tab:learned-baselines}
\small
\setlength{\tabcolsep}{3pt}
\begin{tabular}{lrrr}
\toprule
Method & Natural (\%) $\uparrow$ & Forced (\%) $\uparrow$ & Token count $\downarrow$\\
\midrule
\tableinput{tables/baselines-main.tex}
\bottomrule
\end{tabular}
\end{table}

A separate three-run evaluation checks variation in the base-model reference. With the same model, prompt, sampling, and scoring, it gives 93.44\% natural accuracy and 94.95\% forced accuracy at 4,877 tokens. The main base runs have $+0.45$ natural-accuracy points relative to these runs (paired 95\% interval $[-0.18,+1.10]$) and 1.0\% lower token count.

Table~\ref{tab:deer} varies ten settings of the released DEER procedure \cite{yang_dynamic_2025}, using sampling seed 42 and greedy trial answers capped at 20 tokens. The default combines confidence threshold 0.95, the 10-trial setting, and an 80\% reasoning allowance. Variants change the threshold, use 100 trials, or reduce the allowance to 50\%. The trial parameter limits checks at continuation boundaries; reaching a segment limit or ending without a closing tag can trigger additional checks. All checks count toward token use. The replicated default and 100-trial results use seeds 42--44 with the same model, prompt, policy, and accounting.

\begin{table}[h]
\centering
\caption{DEER settings on Qwen3-4B at 16k. $\theta$ is the confidence threshold, $\rho$ the reasoning-budget fraction, and trials the checking parameter defined above. Accuracy is natural; token counts include checks. Extra counts trial answers, added trial prompts, and discarded generated tokens on MATH-500. Each row reports one run; replicated means appear in Table~\ref{tab:sft}. MATH-500 uses the main overlap exclusions.}
\label{tab:deer}
\small
\setlength{\tabcolsep}{4pt}
\begin{tabular}{rrrrrr}
\toprule
$\theta$ & Trials & $\rho$ & MATH \% & Token count & Extra\\
\midrule
\tableinput{tables/deer.tex}
\bottomrule
\end{tabular}
\end{table}

The lowest-count setting in this sweep reaches 94.33\% accuracy at 2,784 tokens, including its intermediate checks.

We select DEER settings under Settle's development token budgets. We cross $\theta\in\{0.50,0.95\}$ with $\rho\in\{0.50,0.25,0.125\}$ at 100 trials, allowing $\lfloor16{,}384\rho\rfloor$ reasoning tokens and holding other settings fixed. Selection uses 144 development problems, four responses each and seed 42, and maximizes natural accuracy within Settle's mean counts: 5,470.28 for $w=0.25$ (two development runs) and 6,509.89 for $w=0.10$ (three runs). Counts include trial answers, added prompts, and discarded tokens. Ties favor lower count, then a fixed configuration order. The selected $(\theta,\rho)$ settings are $(0.50,0.50)$ for $w=0.10$ and $(0.95,0.25)$ for $w=0.25$. Table~\ref{tab:baseline-development}b gives all six candidates. Both selected policies are evaluated on MATH-500 with seeds 42--44.

Table~\ref{tab:deer-budget-comparisons} reports all four paired comparisons between the selected DEER settings and the two Settle weights.

\begin{table}[!htbp]
\centering
\caption{\textbf{Settle versus development-selected DEER on MATH-500.} Accuracy differences are Settle minus DEER in percentage points, with paired 95\% intervals computed as in Appendix~\ref{app:implementation}. Token savings compare ordinary means, including DEER checks; negative values mean Settle uses more tokens.}
\label{tab:deer-budget-comparisons}
\small
\setlength{\tabcolsep}{5pt}
\begin{tabular}{lrrr}
\toprule
Settle weight & Natural difference & Forced difference & Tokens saved (\%)\\
\midrule
\tableinput{tables/deer-budget-comparisons.tex}
\bottomrule
\end{tabular}
\end{table}

We adapt Halt Vector, the activation-reconstruction procedure of \citet{jayabahu_halt_2026}, to Qwen3-4B. Its compressibility filter retains 48 responses from 24 problems in the shared training pool. We fix steering strength 25 and layer 23, using a proportional-depth layer choice. Attention adapters on layers 0--23 have rank 16 and scaling 32. Training uses 12 epochs, learning rate $10^{-4}$, accumulation over eight responses, reconstruction weight 1, and cosine decay with 10\% warmup. All three fits use these responses and settings, with Settle's base model, training pool, and answer readouts.

Table~\ref{tab:learned-baselines} gives the Qwen3-4B results. On R1-Distill-Qwen-7B, the base model used in the Halt Vector publication, three adapters average 92.30\% natural and 92.42\% forced accuracy at 2,887 tokens. Settle gives 91.03\% natural and 91.62\% forced accuracy at 2,523 tokens under forced scoring; DEER gives 89.36\% natural accuracy at 2,329 tokens including checks.

The closing token can also be encouraged without training. Think Token Adjustment \cite{liu_answer_2025} adds $\alpha(\max_v z_v-\operatorname{mean}_v z_v)$ to its logit until the first closing tag. We evaluate $\alpha\in\{0,0.2,0.4,0.6,0.8,1.0\}$, along with $\alpha=0.6$ using the released answer-format constraints: a newline and boxed-answer prefix after closing. Both $\alpha=0.6$ variants enter the MATH-500 comparison regardless of development rank. No setting meets the 50\%, 60\%, or 75\% development token budget, and plain $\alpha=0.6$ has the highest development natural accuracy, so the selection rule adds no further settings.

Against plain token adjustment, Settle saves 40.4\% of tokens, with a natural-accuracy difference of $+0.84$ points (paired 95\% interval $[-0.08,+1.82]$) and slightly lower forced accuracy (94.73\% versus 94.93\%). This comparison supports an efficiency advantage, with the accuracy ordering depending on the scoring convention. Against the answer-format variant, Settle saves 28.1\% of tokens and gains 7.31 natural-accuracy points $[+5.87,+8.84]$. Tables~\ref{tab:learned-baselines} and~\ref{tab:baseline-development} report both scoring conventions and every development setting.

\begin{table}[H]
\centering
\caption{\textbf{Full development sweeps.} All settings use the same 144 problems, four responses each, and sampling seed 42. SFT and controller evaluations use training seed 0. SFT entries specify learning rate and epoch; every SFT entry uses solutions truncated at their first stable answer. Accuracy is in percent. Panel (a) counts naturally generated tokens; base is $\alpha=0$. Panel (b) includes DEER checks; $\theta$ is the confidence threshold and $\rho$ the reasoning-budget fraction. The final column marks selection under each Settle development ceiling. These problems exclude the training pool.}
\label{tab:baseline-development}
\textit{(a) Training and closing-token baselines}\par\smallskip
\footnotesize
\setlength{\tabcolsep}{4pt}
\begin{tabular}{lrrr}
\toprule
Method and setting & Natural \% & Forced \% & Token count\\
\midrule
\tableinput{tables/baseline-development.tex}
\bottomrule
\end{tabular}
\par\medskip
\textit{(b) DEER at lower reasoning budgets}\par\smallskip
\label{tab:deer-budget-development}
\small
\setlength{\tabcolsep}{5pt}
\begin{tabular}{rrrrl}
\toprule
$\theta$ & $\rho$ & Natural (\%) & Token count & Selected ceiling\\
\midrule
\tableinput{tables/deer-budget-development.tex}
\bottomrule
\end{tabular}
\end{table}
\FloatBarrier
\clearpage
\subsection{PUMA with the same backbone and source traces}
\label{app:puma}
We compare with PUMA's released semantic-redundancy detector and answer-checking procedure \cite{min_stop_2026}, using Qwen3-4B, the shared question prompt, and a 16,384-token cap. The released stopping configuration uses similarity threshold 0.35, confidence threshold 0.98, confidence tolerance 0.03, two consecutive agreeing checks, and a minimum of ten reasoning steps. The implementation replays completed traces to select an exit, then generates the final answer. We count the retained prefix, the generated answer, and all generated tokens and added prompts from checks used before the exit, including checks before the minimum stopping step. Replay supports the accuracy and token-count comparison; the online timing study in Appendix~\ref{app:latency} evaluates base, Settle, and DEER.

For PUMA-SFT, we apply the same exit-labeling procedure to Settle's 1,000 source traces. Following the published filtering rule, we keep correct regenerated answers whose retained reasoning is below 60\% of the source length. This yields 293 training examples. We use the published SFT learning rate $2\times10^{-4}$, three epochs, accumulation over 16 examples, and rank-64 LoRA with scaling 128 on all linear layers. We evaluate the final epoch. This adapts PUMA-SFT to the common backbone and data; the original experiment uses 12,000 problems and R1-Distill-7B.

\begin{table}[H]
\centering
\caption{PUMA comparisons with Qwen3-4B. Accuracy pairs are natural / forced percentages. Four responses per MATH-500 problem, evaluation seed 42, and one training seed per learned method. MATH accuracy excludes the two training overlaps; token means cover all problems and include checks and prompted answers for capped responses. Base uses PUMA's source generations.}
\label{tab:puma}
\footnotesize
\setlength{\tabcolsep}{3pt}
\begin{tabular}{lrr}
\toprule
 & \multicolumn{2}{c}{MATH-500}\\
\cmidrule(lr){2-3}
Method & Accuracy & Tokens\\
\midrule
Base & 93.32 / 94.93 & 4,904\\
PUMA, released replay & 93.42 / 93.67 & 3,433\\
PUMA-SFT, common data & 93.07 / 93.17 & 2,803\\
Settle, $w=0.10$ & 94.18 / 94.58 & 2,890\\
Settle, $w=0.25$ & 93.93 / 93.98 & 2,331\\
\bottomrule
\end{tabular}
\end{table}

On MATH-500, Settle at $w=0.10$ gains 1.10 points in natural accuracy over PUMA-SFT, with paired 95\% problem-bootstrap interval $[-0.05,+2.31]$, at 3.1\% more tokens. Its forced-accuracy gain is 1.41 points $[0.25,2.61]$. Relative to PUMA's inference procedure, the natural-accuracy gain is 0.75 points $[-0.30,+1.86]$ with 15.8\% fewer tokens.

At $w=0.25$, Settle uses 16.8\% fewer tokens than PUMA-SFT, a reduction of 471 tokens with paired 95\% interval $[375,563]$. Its natural-accuracy difference is $+0.85$ points $[-0.20,+1.96]$. Relative to released PUMA replay, it uses 32.1\% fewer tokens, with a natural-accuracy difference of $+0.50$ points $[-0.65,+1.66]$. Both Settle weights are fixed settings from the main comparison.

\FloatBarrier
\section{What the learned stopping score predicts}
\label{app:readouts}
We compare the closing-token score with elapsed-token predictors. The target is stable correctness: the current answer and every later answer are correct.

The native-score and elapsed-token comparisons use identical prefixes, positions, and stable-correct labels: 17,899 boundaries from 498 traces after removing the two training overlaps. The time-only predictor fits logistic regression to standardized cubic-spline features of $\log(1+t)$, with five training-quantile knots, inverse regularization strength 1, and constant extrapolation. It uses 12,000 training boundaries from 300 traces; complexity and regularization are fixed before evaluation. Each native score uses one trained model. Appendix~\ref{app:seeds} identifies the fits used in each analysis. Paired whole-trace bootstrap resamples use seed 0: 5,000 for overall and broad-range AUROC, and 2,000 for exact-token and current-correct analyses, following the convention in Appendix~\ref{app:implementation}.

We compare trained and base closing-token probabilities on the same prefixes and boundary indices; both use only the question and preceding reasoning, without the forced-answer prompt or readout. For $w=0.10$ and $w=0.25$, respectively, paired AUROC gains are $0.299$ $[0.269,0.329]$ and $0.306$ $[0.277,0.336]$ overall, and $0.300$ $[0.266,0.332]$ and $0.311$ $[0.276,0.343]$ at identical token counts. Among currently correct readouts at identical counts, the gains are $0.200$ $[0.112,0.290]$ and $0.210$ $[0.119,0.305]$.

The advantage also holds throughout the trace: native scores exceed the nonlinear time-only predictor in each of the five predefined token ranges (Table~\ref{tab:native-position-strata}a), all of which contain both target classes. Across all boundaries, the paired AUROC improvements are $0.213$ with 95\% interval $[0.176,0.250]$ for $w=0.10$, and $0.221$ with interval $[0.184,0.257]$ for $w=0.25$. This overall AUROC weights boundaries equally.

Pairs compare prefixes from different traces, since stable correctness is monotone within a trace.

Overall AUROC can reflect how far reasoning has progressed. We therefore compare prefixes at matched elapsed lengths. Let $\mathcal P_b$ and $\mathcal N_b$ be stable-correct and other boundaries in token-position group $b$. Conditional AUROC is
\[
\operatorname{AUC}_{\mathrm{cond}}(s)=
\frac{\sum_b\sum_{i\in\mathcal P_b}\sum_{j\in\mathcal N_b}
\left[\ind\{s_i>s_j\}+\tfrac12\ind\{s_i=s_j\}\right]}
{\sum_b|\mathcal P_b|\,|\mathcal N_b|}.
\]
Groups containing both classes contribute in proportion to their number of eligible positive--negative pairs. Table~\ref{tab:native-time-matched}b reports all fixed 64-, 128-, and 256-token groups and exact integer-token matching. Time-only predictions are evaluated once per unique position and reused for all prefixes there, preserving exact ties.

Exact matching includes 12,106 boundaries at 2,315 positions from all 498 traces, forming 16,267 pairs. Each pair compares different traces. Time-only scores tie by construction, while the native scores distinguish stable-correct prefixes at the same elapsed position. Each bootstrap resample recomputes pair weights and the denominator.

\begin{table}[!htbp]
\centering
\caption{\textbf{Native-score prediction conditional on elapsed length.} AUROC [paired whole-trace bootstrap 95\% interval] on 498 nonoverlapping MATH-500 traces. Panel (a) ranks boundaries within each broad range; panel (b) ranks pairs in the same position group. Exact matching admits only identical token counts. Every row uses the same fitted time-only predictor and one agreement-trained model per Settle column. All four matching rules include all 498 traces.}
\label{tab:native-position-strata}
\label{tab:native-time-matched}
\scriptsize
\textit{(a) Ranking within broad token ranges}\par\smallskip
\setlength{\tabcolsep}{3pt}
\begin{tabular}{lrrrr}
\toprule
Elapsed tokens & Boundaries & Time-only predictor & Settle $w=0.10$ & Settle $w=0.25$\\
\midrule
\tableinput{tables/native-position-strata.tex}
\bottomrule
\end{tabular}
\par\medskip
\textit{(b) Ranking matched prefix pairs}\par\smallskip
\setlength{\tabcolsep}{2pt}
\begin{tabular}{lrrrrr}
\toprule
Matching range & Boundaries & Pairs & Time-only predictor & Settle $w=0.10$ & Settle $w=0.25$\\
\midrule
\tableinput{tables/native-time-matched.tex}
\bottomrule
\end{tabular}
\end{table}

To test persistence beyond present correctness, we restrict the comparison to prefixes whose current readout is correct. The subset contains 12,013 boundaries from 457 traces: 11,162 remain correct at every later readout, while 851 boundaries from 121 traces have a later incorrect readout.

Native scores still distinguish persistence within this subset (Table~\ref{tab:current-correct}). Exact-token matching retains 2,455 boundaries from 425 traces at 585 positions, giving 2,015 pairs across traces. For uncertainty, we resample the original 498 traces, apply the correctness restriction, and recompute the statistic; all 2,000 draws contain valid comparisons.

\begin{table}[h]
\centering
\caption{\textbf{Predicting whether a correct answer remains correct.} AUROC [95\% whole-trace bootstrap interval] restricted to currently correct readouts. Overall uses 12,013 boundaries; exact-token matching uses 2,455. $\dagger$ Token-count predictors assign the same score to equal-length prefixes, giving conditional AUROC 0.5.}
\label{tab:current-correct}
\small
\setlength{\tabcolsep}{5pt}
\begin{tabular}{lrr}
\toprule
Score & Overall & Exact token count\\
\midrule
\tableinput{tables/native-current-correct.tex}
\bottomrule
\end{tabular}
\end{table}

\FloatBarrier
\section{Training and evaluation protocol}
\label{app:implementation}
\begin{algorithm}[!htbp]
\caption{Settle: hindsight supervision for the native stop token}
\label{alg:settle}
\begin{algorithmic}[1]
\Require Base policy $\pi_0$, sampled rollouts, checker $E$, weights $w,\lambda$
\For{each recorded rollout $x$}
  \State Find candidate prefix boundaries $t_1<\cdots<t_K$ and terminal point $T$.
  \State Generate short forced readouts $a_1,\ldots,a_K,a_T$.
  \State $r\gets 1$
  \For{$j=K,K-1,\ldots,1$}
    \State $r\gets r\cdot E(a_j,a_T)$; store target $y_j\gets r$.
  \EndFor
  \State Retain supervised boundaries within the training context.
\EndFor
\State Fit a LoRA policy using the stop loss and reference KL in Equation~\ref{eq:loss}.
\State Merge the adapter into the base model.
\Ensure Inference uses ordinary autoregressive decoding and the native \stoptag{} token.
\end{algorithmic}
\end{algorithm}

We consider paragraph breaks and sentence endings followed by \emph{Wait}, \emph{But}, \emph{Alternatively}, \emph{Let me}, or \emph{Hmm}. The first position must follow at least 32 generated tokens, and successive positions must be at least 16 tokens apart. If more than 40 positions remain, we retain at most 40 spaced uniformly through the trace. The terminal readout uses the prefix before the first closing tag, or the full available response when no closing tag occurs. The closing tag \stoptag{} is a single token for Qwen3 and R1-Distill-Qwen-7B. Nemotron splits the tag into multiple tokens; its adaptation supervises the first token, \texttt{</}, and generates the rest through ordinary decoding.

Following budget forcing \cite{muennighoff_s1_2025}, at each position we append \stoptag{}, a \texttt{**Final Answer**} line, and the start of a boxed expression, then generate at most 16 tokens greedily. The same prompt extracts an answer when evaluation reaches the token cap. Ordinary evaluation uses each model's thinking-mode chat template with an instruction to put the final answer in a box. We grade the final boxed expression after the reasoning block with Math-Verify \cite{kydlicek_math-verify_2025}.

We compare answers by their mathematical content. Normalization removes LaTeX delimiters, spacing commands, and other formatting differences; unequal normalized strings are tested for symbolic equivalence with Math-Verify. Parsing failures count as nonmatches. All 40,000 intermediate Qwen3-4B readouts are nonempty. The final readout determines the agreement targets but is excluded from supervised stopping positions.

For the reference-answer comparison, a position is \emph{stable-correct} when its answer and all later readouts are correct, \emph{stable-wrong} when all are incorrect, and \emph{unstable} otherwise. Stable-wrong answers may still change from one incorrect value to another. The two training rules assign the following labels:

\begin{center}
\begin{tabular}{lrr}
\toprule
Reference-answer category & Agreement: stop & Agreement: continue\\
\midrule
Stable-correct & 20,089 & 0\\
Unstable & 0 & 18,264\\
Stable-wrong & 689 & 958\\
\bottomrule
\end{tabular}
\end{center}

The two rules agree at 39,311 of 40,000 positions (98.28\%). Reference-answer supervision assigns stop only to stable-correct positions, whereas agreement supervision also labels 689 consistently wrong positions as stop. These account for 3.32\% of positive targets across 58 traces. Restricting training to the 8,192-token context leaves 39,649 supervised positions. Agreement and stable-correct labels describe the recorded continuation up to the 8,192-token cap, reached by 45.9\% of Qwen3-4B training traces.

The stopping weight determines the penalty for continuing at a positive boundary. We select it by natural accuracy on 144 development problems. Reference-answer supervision considers $w\in\{0.10,0.15,0.25\}$ and selects $0.25$. Agreement supervision considers $\{0.10,0.25\}$ and selects $0.10$. Appendix~\ref{app:seeds} records the development estimates and contributing fits. This agreement weight is carried to the other models without further tuning.

Table~\ref{tab:hparams} gives the Qwen3-4B settings for attention-projection LoRA \cite{hu_lora_2021}, trained on DAPO-Math-17K problems \cite{yu_dapo_2025} with AdamW \cite{loshchilov_decoupled_2019} and a frozen reference model. Training uses only Equation~\ref{eq:loss}: the stopping and KL terms are summed over their respective positions without normalization. Gradients are averaged over eight accumulated examples.

\begin{table}[h]
\centering
\caption{Qwen3-4B training settings.}
\label{tab:hparams}
\begin{tabular}{ll}
\toprule
Item & Setting\\
\midrule
Problem pool & 50 DAPO-Math-17K problems\\
Rollouts & 20 per problem, 1,000 total\\
Rollout sampling & Temperature 1.0, top-$p$ 1.0\\
Generation / training length limit & 8,192 / 8,192 tokens\\
Readout budget & 16 greedy tokens per read\\
LoRA rank / alpha / dropout & 32 / 64 / 0\\
Target projections & Query, key, value, output\\
Epochs / optimizer updates & 3 / 375\\
Batch / gradient accumulation & 1 / 8\\
Optimizer / learning rate & AdamW / $3\times10^{-4}$\\
Weight decay / gradient clip & 0 / 1.0\\
Precision & bfloat16, activation checkpointing\\
Stop loss weight / KL coefficient & 0.10 / 0.20\\
KL direction & $\KL(\pi_0\|\pi_\theta)$\\
Training seeds & 0, 1, 2\\
Deployment & Adapter merged into base weights\\
\bottomrule
\end{tabular}
\end{table}

The learning rate is $3\times10^{-4}$ times
\[
 \min\!\left(1,\frac{s+1}{\max(1,\lfloor0.1U\rfloor)}\right)
 \max\!\left(0,1-\frac{s}{\max(U,1)}\right),
\]
where $s$ is the update step and $U=375$. The schedule combines a warmup factor over the first 10\% of updates with a linear decay factor over all updates.

Qwen3-4B training takes 3,366--3,374 seconds per run on an H100; one training-set readout pass takes 1,922 seconds. These timings exclude data selection, tuning, and evaluation. The collected traces are reused across training runs.

\begin{samepage}
The loss directly changes the competition between closing the reasoning block and producing another token. If $z_t$ is the closing-token logit and $p=p_\theta(t)$, its boundary-loss derivative is
\begin{equation}
 \frac{\partial\ell_j}{\partial z_{t_j}}=(1-y_j)p-wy_j(1-p).
 \label{eq:gradient}
\end{equation}
Stop targets push the closing-token logit upward; continue targets push it downward. The weight $w$ scales the former, and inference uses the ordinary vocabulary softmax.
\end{samepage}

The scoring conventions in Section~\ref{sec:protocol} apply throughout. For forced scoring, every response cut off by the token budget receives an answer prompt, including those whose reasoning block has already closed. Token means include these extra answer tokens unless a table specifies ordinary-generation counts. DEER's counts also include trial answers, their prompts, and discarded generation. Input processing is excluded; Appendix~\ref{app:latency} reports measured inference time.

MATH-500 accuracy excludes problems 292 and 475 because they overlap the training data; token means use all 500 problems. MATH-500 uses four responses per problem.

We first average responses within each problem, also averaging across runs for replicated comparisons, then draw 20,000 paired problem bootstrap samples using NumPy's default generator with seed 0. The 2.5th and 97.5th percentiles give pointwise 95\% intervals without adjustment for multiple comparisons. These intervals describe variation across problems for the evaluated fits; single-fit comparisons use the same procedure and do not estimate training-run variability. The diagnostics resample whole traces, with counts specified in Appendix~\ref{app:readouts}.
\FloatBarrier
\section{When answers stabilize}
\label{app:census}
Early stable answers define an opportunity for learned stopping. Table~\ref{tab:census} measures that opportunity in separately sampled base-model traces: a retrospective rule chooses the first position whose answer remains correct at every later readout. When there is no such position, it uses the final answer, which may be obtained by a forced readout. The resulting accuracy and token counts describe what can be recovered from these collected traces.

\begin{table}[h]
\centering
\caption{When answers stabilize in sampled reasoning traces. Settle is the percentage with an early stable-correct answer; median is the first such position as a fraction of trace length among those that settle. Accuracy and trace-token pairs are ordinary completion / retrospective stopping. ``Repeat'' denotes a separate Qwen3-4B MATH-500 sample.}
\label{tab:census}
\scriptsize
\setlength{\tabcolsep}{3pt}
\begin{tabular}{lrrrrr}
\toprule
Evaluation & Traces & Settle \% & Median & Accuracy \% & Trace tokens\\
\midrule
\tableinput{tables/census.tex}
\bottomrule
\end{tabular}
\end{table}

\FloatBarrier
\section{Evaluation on separate MATH problems}
\label{app:chronology}
On 500 further test problems, Settle reduces mean token count by 41.5\% at 8k and 42.7\% at 16k, with small positive mean changes in forced accuracy (Table~\ref{tab:math-sample}). This sample from MATH \cite{hendrycks_math_2021} excludes the training and development problems and all MATH problems already included in the benchmark evaluations.

\begin{table}[h]
\centering
\caption{Separate 500-problem MATH test sample, Qwen3-4B. Forced accuracy, paired 95\% intervals, and base / Settle token means, averaged over three runs.}
\label{tab:math-sample}
\footnotesize
\setlength{\tabcolsep}{4pt}
\begin{tabular}{lrrlrr}
\toprule
Cap & Base & Settle & Change [95\% interval] & Token count & Saved \%\\
\midrule
\tableinput{tables/math-sample.tex}
\bottomrule
\end{tabular}
\end{table}

At 8k, natural accuracy rises by 4.08 points, indicating that earlier completion contributes to the larger natural-scoring gain.
\FloatBarrier
\section{Accuracy and token counts across models and tasks}
\label{app:transfer}
We next examine how the accuracy--token-count tradeoff changes across models and tasks while keeping the stopping weight at $w=0.10$. Table~\ref{tab:models} reports all four models on MATH-500 under both natural and forced scoring. Nemotron uses the first-token adaptation in Appendix~\ref{app:implementation}; the other models have a single-token closing tag.

\begin{table}[h]
\centering
\caption{Results across four models. Accuracy pairs are base / \method{}, in percent; forced-accuracy changes are percentage points. Each model uses three base runs and three trained runs. Token savings include additional answer tokens under forced scoring.}
\label{tab:models}
\scriptsize
\setlength{\tabcolsep}{2pt}
\begin{tabular}{llrrlr}
\toprule
Model & Set & Natural & Forced & Forced change [95\% CI] & Saved \%\\
\midrule
\tableinput{tables/models.tex}
\bottomrule
\end{tabular}
\end{table}

With the same stopping weight across tasks, Settle reduces mean token count on every benchmark in Table~\ref{tab:suite}, with savings from 13.2\% to 49.1\%. The table gives the accompanying accuracy changes. The suite covers GSM8K \cite{cobbe_training_2021}; AMC23 \cite{mathematical_association_of_america_american_nodate}; Minerva \cite{lewkowycz_solving_2022}; OlympiadBench \cite{he_olympiadbench_2024}; and GPQA-Diamond \cite{rein_gpqa_2023}. Each comparison uses one base run and three trained models with a common evaluation seed. Problem and response counts differ from the main evaluation and are listed in the table. The accompanying MATH-500 evaluation uses one response per problem, includes both training overlaps, and gives 95.2\% / 94.9\% base / Settle forced accuracy.

\begin{table}[h]
\centering
\caption{Additional benchmarks, Qwen3-4B at 16k. Forced accuracy and token pairs are base / \method{}. $N$ denotes problems and $S$ responses per problem. Results average three trained models against one base run.}
\label{tab:suite}
\footnotesize
\setlength{\tabcolsep}{3pt}
\begin{tabular}{lrrrrrr}
\toprule
Set & $N$ & $S$ & Accuracy \% & Change & Token count & Saved \%\\
\midrule
\tableinput{tables/suite.tex}
\bottomrule
\end{tabular}
\end{table}
\FloatBarrier
\subsection{Calibrating stopping pressure on harder problems}
\label{app:calibration}
We train Qwen3-4B at $w\in\{0.01,0.025,0.05,0.10\}$ on the original 1,000 traces, keeping $\lambda=0.2$, the 375 training updates, and the remaining settings fixed. Each weight uses training seed 0. We choose the weight on 128 harder DAPO development problems, excluded from training and the evaluation benchmarks. The rule selects the lowest token count among settings whose forced accuracy is within one percentage point of the base model. Four responses per development problem select $w=0.025$; the test comparisons below use this fixed development-selected weight.

\begin{table}[H]
\centering
\caption{Development accuracy for Qwen3-4B stopping-pressure calibration. One training seed per weight and four responses per development problem, with evaluation seed 42 and a 16k cap. $\dagger$ The development-selected weight.}
\label{tab:hard-calibration}
\footnotesize
\setlength{\tabcolsep}{3pt}
\begin{tabular}{lr}
\toprule
Method & Forced accuracy (\%)\\
\midrule
Base & 85.55\\
Settle, $w=0.01$ & 86.13\\
Settle, $w=0.025^\dagger$ & 84.57\\
Settle, $w=0.05$ & 84.18\\
Settle, $w=0.10$ & 80.47\\
\bottomrule
\end{tabular}
\end{table}

\begin{table}[H]
\centering
\caption{The same Qwen3-4B weight sweep on MATH-500 and AMC23. Accuracy pairs are natural / forced percentages. Four responses per MATH problem and eight per AMC problem, evaluation seed 42, and one training seed per weight. MATH base uses the source generations from Table~\ref{tab:puma}; all rows use the shared prompts, 16k cap, and scoring. $\dagger$ The weight selected on hard development problems.}
\label{tab:calibration-math-amc}
\small
\begin{tabular}{lrrrr}
\toprule
 & \multicolumn{2}{c}{MATH-500} & \multicolumn{2}{c}{AMC23}\\
\cmidrule(lr){2-3}\cmidrule(lr){4-5}
Method & Accuracy & Token count & Accuracy & Token count\\
\midrule
\tableinput{tables/calibration-math-amc.tex}
\bottomrule
\end{tabular}
\end{table}

On AMC23, the development-selected $w=0.025$ reduces tokens by 8.1\%, with a forced-accuracy difference of $-0.94$ points $[-3.44,+0.94]$. The token reduction is 611, with paired 95\% interval $[338,914]$.

\subsection{Training on harder problems and reference-answer labels}
\label{app:training-grid}
We cross two training pools with agreement or reference-answer labels, keeping $w=0.10$, $\lambda=0.2$, 375 updates, and the remaining training settings fixed. Both pools contain 1,000 traces. The original pool uses 20 traces on each of 50 problems. The broader pool retains ten traces per original problem and adds two traces on each of 250 harder DAPO problems. We sample the added problems from a preliminary eight-response evaluation of 5,000 DAPO questions, retaining those with forced accuracy in $(0,0.5]$ and excluding training, development, and benchmark overlaps. Qwen3-4B achieves 30.8\% natural accuracy on the added traces under the 8k generation cap, compared with 54.0\% on the original pool.

Reference-answer targets require every subsequent readout to be correct; all other boundaries receive continue targets. Agreement targets instead require agreement with the terminal readout. Within each pool, both conditions use identical traces, sampled positions, and KL masks. After the training-context limit, the label rules differ at 632 of 39,649 original-pool boundaries and 1,915 of 39,553 broader-pool boundaries. We train all four conditions with seeds 0--2, paired with evaluation seeds 42--44.

\begin{table}[H]
\centering
\caption{Training-pool and label-rule comparison on Qwen3-4B, with $w=0.10$ throughout. Means $\pm$ sample standard deviations across three training runs. Four responses per MATH problem, with a 16k cap. Token counts include prompted answers for capped responses.}
\label{tab:training-grid}
\small
\setlength{\tabcolsep}{5pt}
\begin{tabular}{llrrr}
\toprule
Pool & Labels & Natural accuracy (\%) & Forced accuracy (\%) & Token count\\
\midrule
\tableinput{tables/training-grid.tex}
\bottomrule
\end{tabular}
\end{table}

\FloatBarrier
\section{Inference time under batching}
\label{app:latency}
We measure inference time on one H100 80GB using vLLM 0.28 \cite{kwon_efficient_2023}, with 100 MATH problems, one response per problem, and a 16k cap. Each setting is timed once after model loading. Settle uses the Qwen3-4B model trained with seed 1. Base and Settle use ordinary generation; DEER uses the procedure in Appendix~\ref{app:baselines}. Batch sizes 32 and 100 process fixed groups, with no new requests arriving during generation.

\begin{table}[H]
\centering
\caption{Inference time for 100 MATH problems on one H100. Total time covers all requests; median time per request is reported only for batch size 1. Accuracy is natural accuracy on these requests. Token counts here cover responses only; the accuracy--cost comparisons above also count the additional work of DEER's intermediate checks.}
\label{tab:latency}
\footnotesize
\setlength{\tabcolsep}{5pt}
\begin{tabular}{lrrrrr}
\toprule
Method & Batch & Total time (s) & Median (s) & Accuracy \% & Response tokens\\
\midrule
\tableinput{tables/latency.tex}
\bottomrule
\end{tabular}
\end{table}

At batch size 1, Settle reduces median request time from 13.25 to 7.66 seconds; DEER's median is approximately 7 seconds at coarser timing resolution. For batches of 32 and 100, Settle completes the workload 1.72 and 1.86 times as fast as DEER. Table~\ref{tab:latency} gives accuracy alongside time for each setting. Its token column covers final responses; the policy comparisons additionally count DEER's intermediate checks.
\FloatBarrier
\section{Results across individual runs}
\label{app:seeds}
All tables in this section report individual runs at a 16,384-token response cap. MATH-500 accuracy uses 498 nonoverlapping problems; token means cover all 500. Inference-only methods use sampling seeds 42--44. Learned methods pair training seeds 0--2 with sampling seeds 42--44. For ordinary generation, the MATH-500 evaluator generates four responses per prompt with child seeds $r,\ldots,r+3$ for evaluation seed $r$, so runs 42--44 use overlapping seed ranges. DEER derives separate seeds for each sample and decoding stage. All responses from a problem remain together in the bootstrap.

\paragraph{Run accounting.} The $w=0.10$ main evaluation uses the second seed-0 fit plus seeds 1 and 2; the native-score analysis uses the first seed-0 fit with identical settings. Substituting the first fit changes three-run mean natural/forced accuracy by $+0.12/+0.30$ points at 8k and $-0.05/-0.07$ at 16k, with token changes below 0.3\% at both caps. The $w=0.25$ results use one fit per seed. In development selection, agreement weights $0.10$ and $0.25$ reach approximately 80.3\% and 79.0\% natural accuracy; the latter estimate averages two trained models. One regularizer-removal evaluation used disjoint batches across three GPUs with the same per-response seeds.


\begingroup
\fontsize{9.04}{10.5}\selectfont
\setlength{\tabcolsep}{5pt}
\setlength{\LTcapwidth}{\linewidth}
\setlength{\LTpre}{8pt}
\setlength{\LTpost}{8pt}
\begin{longtable}{lrrr}
\caption{\textbf{Individual runs across models.} Accuracy pairs are natural / forced; token means include additional answer tokens under forced scoring.}\label{tab:perseed}\\
\toprule
Method & Seed & \multicolumn{2}{c}{MATH-500}\\
\cmidrule(lr){3-4}
& & Accuracy (\%) & Token count\\
\midrule
\endfirsthead
\caption[]{Individual runs across models (continued)}\\
\toprule
Method & Seed & \multicolumn{2}{c}{MATH-500}\\
\cmidrule(lr){3-4}
& & Accuracy (\%) & Token count\\
\midrule
\endhead
\midrule
\multicolumn{4}{r}{\textit{Continued on next page}}\\
\endfoot
\bottomrule
\endlastfoot
\addlinespace[3pt]
\multicolumn{4}{l}{\textit{Qwen3-4B}}\\*
Base & 42 & 94.33 / 95.28 & 4,816\\*
Base & 43 & 94.03 / 95.43 & 4,809\\*
Base & 44 & 93.32 / 94.98 & 4,862\\*
Settle & 0 & 94.63 / 94.88 & 2,847\\*
Settle & 1 & 94.63 / 94.88 & 3,004\\*
Settle & 2 & 93.88 / 94.43 & 2,861\\
\addlinespace[3pt]
\multicolumn{4}{l}{\textit{Qwen3-8B}}\\*
Base & 42 & 93.32 / 95.53 & 5,102\\*
Base & 43 & 93.93 / 95.73 & 5,056\\*
Base & 44 & 93.88 / 95.73 & 5,177\\*
Settle & 0 & 94.68 / 95.58 & 3,201\\*
Settle & 1 & 94.18 / 95.03 & 3,055\\*
Settle & 2 & 95.13 / 95.73 & 3,150\\
\addlinespace[3pt]
\multicolumn{4}{l}{\textit{Nemotron-Nano-8B}}\\*
Base & 42 & 94.63 / 94.88 & 3,227\\*
Base & 43 & 94.78 / 94.88 & 3,234\\*
Base & 44 & 94.43 / 94.68 & 3,242\\*
Settle & 0 & 92.67 / 92.57 & 2,191\\*
Settle & 1 & 93.07 / 93.22 & 2,153\\*
Settle & 2 & 93.22 / 93.37 & 2,128\\
\addlinespace[3pt]
\multicolumn{4}{l}{\textit{R1-Distill-Qwen-7B}}\\*
Base & 42 & 92.87 / 93.83 & 3,743\\*
Base & 43 & 92.37 / 92.97 & 3,659\\*
Base & 44 & 93.22 / 93.88 & 3,657\\*
Settle & 0 & 91.16 / 91.77 & 2,611\\*
Settle & 1 & 90.81 / 91.47 & 2,518\\*
Settle & 2 & 91.11 / 91.62 & 2,440\\
\end{longtable}
\endgroup

\begingroup
\fontsize{9.04}{10.5}\selectfont
\setlength{\tabcolsep}{5pt}
\setlength{\LTcapwidth}{\linewidth}
\setlength{\LTpre}{8pt}
\setlength{\LTpost}{8pt}
\begin{longtable}{p{2.6in}rrrr}
\caption{\textbf{Individual stopping-baseline runs.} Token counts cover ordinary generation. Epoch-2 unregularized truncated SFT appears in Table~\ref{tab:anchor-perseed}, with both token conventions.}\label{tab:baseline-perseed}\\
\toprule
Method and setting & Run & Natural (\%) & Forced (\%) & Tokens\\
\midrule
\endfirsthead
\caption[]{Individual stopping-baseline runs (continued)}\\
\toprule
Method and setting & Run & Natural (\%) & Forced (\%) & Tokens\\
\midrule
\endhead
\midrule
\multicolumn{5}{r}{\textit{Continued on next page}}\\
\endfoot
\bottomrule
\endlastfoot
\addlinespace[3pt]
\textit{Base (repeat evaluation)} & 0 & 93.22 & 94.83 & 4,905\\*
 & 1 & 93.78 & 95.23 & 4,853\\*
 & 2 & 93.32 & 94.78 & 4,874\\
\addlinespace[3pt]
\textit{Halt Vector} & 0 & 93.78 & 94.53 & 4,116\\*
 & 1 & 93.83 & 94.63 & 4,137\\*
 & 2 & 93.93 & 94.68 & 4,244\\
\addlinespace[3pt]
\textit{LSTM controller, $\tau=0.95$} & 0 & 84.44 & 84.44 & 2,859\\*
 & 1 & 88.45 & 88.96 & 3,457\\*
 & 2 & 88.96 & 89.66 & 3,591\\
\addlinespace[3pt]
\textit{LSTM controller, $\tau=0.99$} & 0 & 87.75 & 87.80 & 3,484\\*
 & 1 & 92.47 & 93.47 & 4,392\\*
 & 2 & 91.37 & 92.42 & 4,237\\
\addlinespace[3pt]
\textit{LSTM controller, $\tau=0.995$} & 0 & 89.11 & 89.26 & 3,767\\*
 & 1 & 93.12 & 94.33 & 4,595\\*
 & 2 & 91.82 & 92.92 & 4,424\\
\addlinespace[3pt]
\textit{LSTM controller, $\tau=0.9999$} & 0 & 93.17 & 94.33 & 4,730\\*
 & 1 & 93.78 & 95.23 & 4,853\\*
 & 2 & 93.32 & 94.78 & 4,868\\
\addlinespace[3pt]
\textit{SFT: stability-truncated, $3\!\times\!10^{-4}$, epoch 1} & 0 & 63.81 & 63.81 & 1,544\\*
 & 1 & 88.20 & 88.40 & 2,488\\*
 & 2 & 88.91 & 89.11 & 2,718\\
\addlinespace[3pt]
\textit{Think Token Adjustment, $\alpha=0.6$} & 0 & 93.52 & 95.08 & 4,900\\*
 & 1 & 93.72 & 94.98 & 4,892\\*
 & 2 & 93.37 & 94.73 & 4,824\\
\addlinespace[3pt]
\textit{Think Token Adjustment, $\alpha=0.6$ (answer format)} & 0 & 86.14 & 86.80 & 3,988\\*
 & 1 & 86.09 & 86.80 & 3,934\\*
 & 2 & 88.96 & 90.06 & 4,185\\
\end{longtable}
\endgroup

\begingroup
\fontsize{9.04}{10.5}\selectfont
\setlength{\tabcolsep}{5pt}
\setlength{\LTcapwidth}{\linewidth}
\setlength{\LTpre}{8pt}
\setlength{\LTpost}{8pt}
\begin{longtable}{rrrrr}
\caption{\textbf{Individual reference-regularization and DEER runs.} Ordinary token counts precede additional answer forcing; the final column includes it. DEER counts include intermediate checks.}\label{tab:anchor-perseed}\label{tab:deer-budget-perseed}\\
\toprule
Sampling seed & Natural (\%) & Forced (\%) & \multicolumn{2}{c}{Token count}\\
\cmidrule(lr){4-5}
& & & Ordinary & With forced answers\\
\midrule
\endfirsthead
\caption[]{Individual reference-regularization and DEER runs (continued)}\\
\toprule
Sampling seed & Natural (\%) & Forced (\%) & \multicolumn{2}{c}{Token count}\\
\cmidrule(lr){4-5}
& & & Ordinary & With forced answers\\
\midrule
\endhead
\midrule
\multicolumn{5}{r}{\textit{Continued on next page}}\\
\endfoot
\bottomrule
\endlastfoot
\addlinespace[3pt]
\multicolumn{5}{l}{\textit{Truncated SFT, $\beta=0$}}\\*
42 & 89.36 & 89.41 & 2,347.7 & 2,347.8\\*
43 & 84.54 & 84.59 & 2,268.7 & 2,268.8\\*
44 & 88.45 & 88.60 & 2,419.5 & 2,419.6\\
\addlinespace[3pt]
\multicolumn{5}{l}{\textit{Truncated SFT, $\beta=0.2$}}\\*
42 & 89.81 & 90.06 & 2,591.2 & 2,591.3\\*
43 & 85.39 & 85.64 & 2,460.7 & 2,460.8\\*
44 & 89.56 & 89.86 & 2,639.1 & 2,639.2\\
\addlinespace[3pt]
\multicolumn{5}{l}{\textit{Truncated SFT, $\beta=1.0$}}\\*
42 & 91.67 & 92.12 & 3,258.2 & 3,258.3\\*
43 & 88.25 & 88.86 & 3,014.6 & 3,014.8\\*
44 & 91.57 & 91.92 & 3,116.3 & 3,116.5\\
\addlinespace[3pt]
\multicolumn{5}{l}{\textit{Truncated SFT, $\beta=5.0$}}\\*
42 & 93.52 & 94.58 & 4,018.3 & 4,018.6\\*
43 & 92.72 & 93.57 & 3,851.1 & 3,851.3\\*
44 & 92.62 & 93.42 & 3,784.0 & 3,784.2\\
\addlinespace[3pt]
\multicolumn{5}{l}{\textit{Settle, $w=0.25$, $\lambda=0$}}\\*
42 & 0.00 & 0.55 & 1,353.2 & 1,353.9\\*
43 & 0.00 & 2.71 & 13,840.2 & 13,845.4\\*
44 & 0.10 & 1.51 & 3,672.5 & 3,675.3\\
\addlinespace[3pt]
\multicolumn{5}{l}{\textit{Settle, $w=0.25$, $\lambda=0.2$}}\\*
42 & 93.52 & 93.62 & 2,303.3 & 2,303.4\\*
43 & 93.88 & 93.98 & 2,423.3 & 2,423.4\\*
44 & 93.42 & 93.52 & 2,339.7 & 2,339.8\\
\addlinespace[3pt]
\multicolumn{5}{l}{\textit{DEER, default}}\\*
42 & 94.68 & 94.68 & 3,641.5 & 3,641.6\\*
43 & 94.98 & 94.98 & 3,661.4 & 3,661.4\\*
44 & 94.78 & 94.78 & 3,655.1 & 3,655.2\\
\addlinespace[3pt]
\multicolumn{5}{l}{\textit{DEER, 100 trials}}\\*
42 & 94.88 & 94.88 & 3,131.9 & 3,131.9\\*
43 & 94.83 & 94.83 & 3,133.8 & 3,133.8\\*
44 & 94.73 & 94.73 & 3,193.6 & 3,193.7\\
\addlinespace[3pt]
\multicolumn{5}{l}{\textit{DEER, $\theta=0.50, \rho=0.5$}}\\*
42 & 94.03 & 94.08 & 2,805.5 & 2,805.5\\*
43 & 94.03 & 94.03 & 2,762.9 & 2,762.9\\*
44 & 93.83 & 93.93 & 2,817.2 & 2,817.3\\
\addlinespace[3pt]
\multicolumn{5}{l}{\textit{DEER, $\theta=0.95, \rho=0.25$}}\\*
42 & 92.82 & 92.82 & 2,650.7 & 2,650.7\\*
43 & 92.82 & 92.92 & 2,647.4 & 2,647.5\\*
44 & 92.52 & 92.52 & 2,674.6 & 2,674.6\\
\end{longtable}
\endgroup

\begin{table}[H]
\centering
\caption{\textbf{Individual Settle reference-weight sensitivity runs.} Qwen3-4B, $w=0.25$. These twelve fits use the same source traces and training settings as the main models, including three new runs at $\lambda=0.20$. Token counts cover ordinary generation.}
\label{tab:lambda-perseed}
\small
\setlength{\tabcolsep}{6pt}
\begin{tabular}{rrrrrr}
\toprule
$\lambda$ & Training seed & Sampling seed & Natural (\%) & Forced (\%) & Token count\\
\midrule
\tableinput{tables/lambda-perseed.tex}
\bottomrule
\end{tabular}
\end{table}

\end{document}